\documentclass[10pt,twocolumn,letterpaper]{article}

\usepackage[pagenumbers]{cvpr} % To force page numbers, e.g. for an arXiv version

\usepackage{xspace}

\newcommand{\hygeo}{\textup{\textbf{\texttt{HyGeoBench}}}\xspace}
\newcommand{\hygeoagent}{\textup{\textbf{\texttt{HyGeoAgent}}}\xspace}
\usepackage{array}
\usepackage{tabularx}
\usepackage{multirow}
\usepackage{tikz}
\usetikzlibrary{arrows.meta,fit}
\usepackage{placeins}

\definecolor{cvprblue}{rgb}{0.21,0.49,0.74}
\usepackage[pagebackref,breaklinks,colorlinks,allcolors=cvprblue]{hyperref}

\def\confName{CVPR}
\def\confYear{2027}

\title{Benchmarking Reasoning Reliability, Prior–Evidence Contrast, and Spectral Discrimination in Remote-Sensing Geological Image Interpretation}

\author{
\textbf{Jun Wang}$^{1,*}$,\quad
\textbf{Fengpeng Li}$^{2,*}$,\quad
\textbf{Wei Han}$^{1,*}$,\quad
\textbf{Yuxuan Li}$^{1}$,\quad
\textbf{Tianjin Huang}$^{3}$,\\
\textbf{Jining Yan}$^{1}$,\quad
\textbf{Sheng Wang}$^{1}$,\quad
\textbf{Kan Sun}$^{4}$,\quad
\textbf{Lizhe Wang}$^{5,\dagger}$\\[0.6em]
$^{1}$School of Computer Science and Artificial Intelligence, China University of Geosciences\\
$^{2}$PRADA Lab, King Abdullah University of Science and Technology\\
$^{3}$Department of Computer Science, University of Exeter\\
$^{4}$School of Mathematics and Computer Science, Wuhan Polytechnic University\\
$^{5}$Tsinghua Shenzhen International Graduate School, Tsinghua University
}

\makeatletter
\apptocmd{\@maketitle}{%
    \par\noindent
    \begin{minipage}{\textwidth}
        \centering
        \captionsetup{hypcap=false}
        \includegraphics[width=\linewidth]{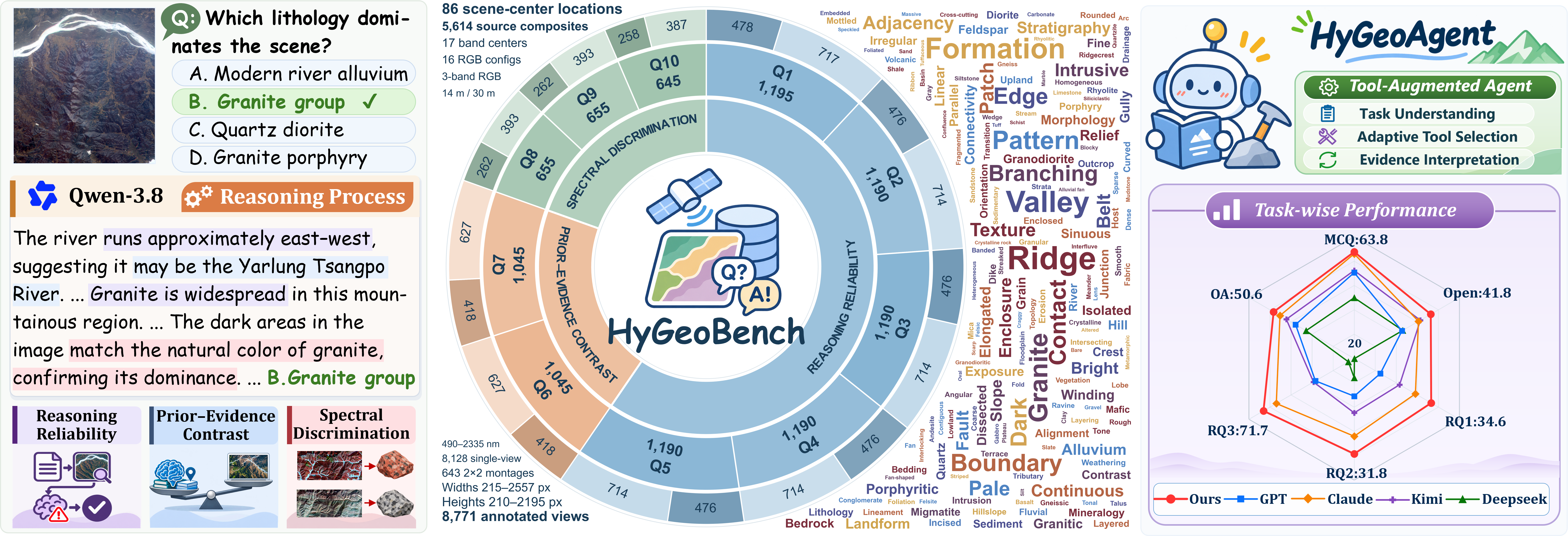}
        \captionof{figure}{Overview of \hygeo and \hygeoagent for evaluating and improving geological image interpretation.}
        \label{fig:case_comparison}
    \end{minipage}
    \par\vspace{12pt} % 图1与下方正文的间距
}{}{\PackageError{hygeo}{Could not attach Figure 1 to the title block}{Check the definition of \string\@maketitle.}}
\makeatother

\begin{document}
\maketitle
% Match the existing author markers without changing body footnote numbering.
% \begingroup
%     \renewcommand{\thefootnote}{\fnsymbol{footnote}}
%     \footnotetext[1]{Equal contribution.}
%     \footnotetext[2]{Corresponding author (\texttt{lizhe.wang@sz.tsinghua.edu.cn}).}
% \endgroup

\begin{abstract}

Reliable geological image interpretation requires accurate geological recognition and reasoning grounded in visual evidence. Although Multimodal Large Language Models show promise in remote sensing, their reliability in geological interpretation remains unclear. We introduce \hygeo, a benchmark of \textbf{10,000 questions} based on hyperspectral imagery across three dimensions: reasoning reliability, prior--evidence contrast, and spectral discrimination. It covers ten task types, with both multiple-choice and open-ended response formats. Integrating hyperspectral satellite imagery with geological annotations and verified evidence, \hygeo examines reasoning errors, conflicts between geological priors and visual evidence, and cross-band inconsistencies. Evaluations reveal limitations in fine-grained recognition and evidence-grounded interpretation. To address these challenges, we propose \hygeoagent, a tool-augmented framework integrating region observation, text knowledge retrieval, same-band image retrieval, and native PCA analysis through adaptive tool use. Experiments demonstrate consistent improvements over the corresponding base models, improving geological image interpretation across diagnostic dimensions.

\end{abstract}

\section{Introduction}
\label{sec:intro}

Remote sensing (RS) geological image interpretation supports regional geological surveys, lithological mapping, and mineral resource exploration~\cite{vandermeer2012geologic,cudahy2016mineralmapping}. Hyperspectral remote sensing is particularly valuable for geological interpretation, as its rich spectral information helps distinguish lithologies and characterize mineral compositions~\cite{peyghambari2021hyperspectral,hajaj2024hyperspectral}. Geological interpretation integrates surface morphology, spatial distributions, spectral responses, colors, textures, and geological knowledge to infer geological units, lithological compositions, and structural characteristics~\cite{richards2022remote,gupta2018geology}. Unlike conventional object recognition, geological interpretation requires linking visual observations to geological properties, identifying targets, and determining which conclusions are supported by available evidence. Reliable interpretation therefore depends on both accurate visual perception and effective integration of domain knowledge~\cite{stacot2025}.

Multimodal Large Language Models (MLLMs) combine visual representations with language modeling and instruction following~\cite{li2023blip2,liu2023llava}. Domain-specific models extend these capabilities to remote sensing image understanding~\cite{rsgpt2025,geochat2024,lhrsbot2024,earthgpt2024}, supporting landform description, lithology comparison, and multi-step geological reasoning~\cite{stacot2025}. However, studies have identified object hallucinations and failures to resolve conflicts between language priors and visual evidence in multimodal models~\cite{li2023pope,guan2024hallusionbench}. Errors can also occur on image pairs readily distinguishable to humans~\cite{tong2024eyeswideshut}. In geological interpretation, models may rely on inaccurate visual observations or invoke knowledge without grounding it in specific targets~\cite{stacot2025}. Thus, correct final answers alone do not guarantee reliable reasoning~\cite{turpin2023unfaithful}.

Figure~\ref{fig:case_comparison} illustrates the motivation of our study and the overall frameworks of \hygeo and \hygeoagent. In the left panel, Qwen-3.8 Max~\cite{qwen2026qwen38max} correctly identifies the dominant lithology as a granite group, but its reasoning process is flawed. It incorrectly associates an east--west river with the Yarlung Tsangpo River and interprets dark image regions as evidence of granite dominance, neither of which is supported by the available visual or geological evidence. Thus, despite producing the correct final answer, the model reaches its conclusion through erroneous and unsupported reasoning. This example highlights a critical limitation: answer correctness alone does not guarantee reliable geological understanding, and generated explanations may not faithfully reflect the underlying decision process~\cite{turpin2023unfaithful}. These observations motivate three research questions: (1) \textbf{Reasoning Reliability:} Where do errors occur across geological interpretation stages, and how do they affect final judgments? (2) \textbf{Prior--Evidence Contrast:} How do models reconcile general geological knowledge with scene-specific visual evidence when the two lead to conflicting interpretations? (3) \textbf{Spectral Discrimination:} Can models discriminate band-induced appearance variations from genuine geological differences across spectral views?

We develop \hygeo for evaluating geological reasoning and \hygeoagent for improving geological image interpretation. Our main contributions are:

\begin{itemize}
    
    \item \textbf{\hygeo: A benchmark for geological interpretation of hyperspectral remote sensing imagery.}
    We construct a benchmark covering three dimensions: reasoning reliability, prior--evidence contrast, and spectral discrimination, to assess hierarchical geological reasoning, aligning geological knowledge with scene-specific visual evidence, and distinguishing band-induced appearance variations from geological differences.

    \item \textbf{Systematic evaluation of multimodal models and their limitations.}
    We evaluate approximately 20 representative open- and closed-source MLLMs and agents, revealing unreliable reasoning, geological prior dominance over visual evidence, and confusion of band-dependent appearance variations with geological changes.
    
    \item \textbf{\hygeoagent: A tool-augmented agent for geological image interpretation.}
    We integrate region observation, text knowledge retrieval, same-band image retrieval, and native PCA analysis within a single-MLLM framework. Through adaptive tool use, \hygeoagent improves geological interpretation through tool-assisted integration of geological knowledge and visual evidence.
    
\end{itemize}

\section{Related Work}
\label{sec:relatedwork}

\textbf{Remote Sensing VQA Benchmarks.}
Remote sensing VQA benchmarks have evolved toward diverse tasks and complex reasoning. EarthVQA~\cite{earthvqa2024} explored object relations, MMRS-1M~\cite{earthgpt2024} integrated multisensor understanding, HRVQA~\cite{hrvqa2024} focused on high-resolution imagery, VRSBench~\cite{vrsbench2024} covered diverse vision--language tasks, and LHRS-Bench~\cite{lhrsbot2024} introduced hierarchical evaluation. Subsequent benchmarks, including RSIEval~\cite{rsgpt2025}, XLRS-Bench~\cite{xlrsbench2025}, GeoMap-Bench~\cite{peace2025}, LRS-VQA~\cite{lrsvqa2025}, and CHOICE~\cite{choice2025}, further explored open-ended reasoning, ultra-high-resolution imagery, geological maps, and hierarchical perception. FRIEDA~\cite{frieda2026} and GeoMMBench~\cite{geommbench2026} extended evaluation to spatial reasoning and expert-level geoscientific understanding. However, fine-grained lithological discrimination, prior--evidence conflicts, and cross-band consistency remain underexplored. We introduce \hygeo to address these gaps through diagnostic evaluation of geological reasoning.

\textbf{Remote Sensing MLLMs.}
Remote sensing MLLMs have progressed from image--text representation learning toward multisource understanding and spatial reasoning. RemoteCLIP~\cite{remoteclip2024}, EarthGPT~\cite{earthgpt2024}, GeoChat~\cite{geochat2024}, and LHRS-Bot~\cite{lhrsbot2024} explored visual representation, multisensor understanding, spatial grounding, and hierarchical alignment. Subsequent models, including RSGPT~\cite{rsgpt2025}, SkyEyeGPT~\cite{skyeyegpt2025}, TEOChat~\cite{teochat2025}, EarthDial~\cite{earthdial2025}, GeoPixel~\cite{geopixel2025}, STA-CoT~\cite{stacot2025}, and GeoLLaVA-8K~\cite{geollava8k2025}, advanced domain adaptation, multilevel understanding, temporal and multispectral analysis, pixel-level grounding, and geological reasoning. Earth-Agent~\cite{earthagent2026} and TerraScope~\cite{terrascope2026} further explored tool-assisted multisource analysis and pixel-level spatial reasoning. However, grounding geological knowledge in visual evidence and distinguishing spectral from geological variations remain challenging. We introduce \hygeoagent for evidence-grounded geological interpretation.
% HyGeoBench — revised CVPR main-text version.
% Original mathematical equations and all figure/table environments are preserved.
% The five original tables are grouped AFTER the chapter body, before any \end{document}.
% Requires: graphicx, booktabs, tabularx, array, multirow, amsmath.
% Uses the existing \hygeo macro and bibliography entries.

\begin{figure*}[t]
    \centering
    \includegraphics[width=0.95\textwidth]{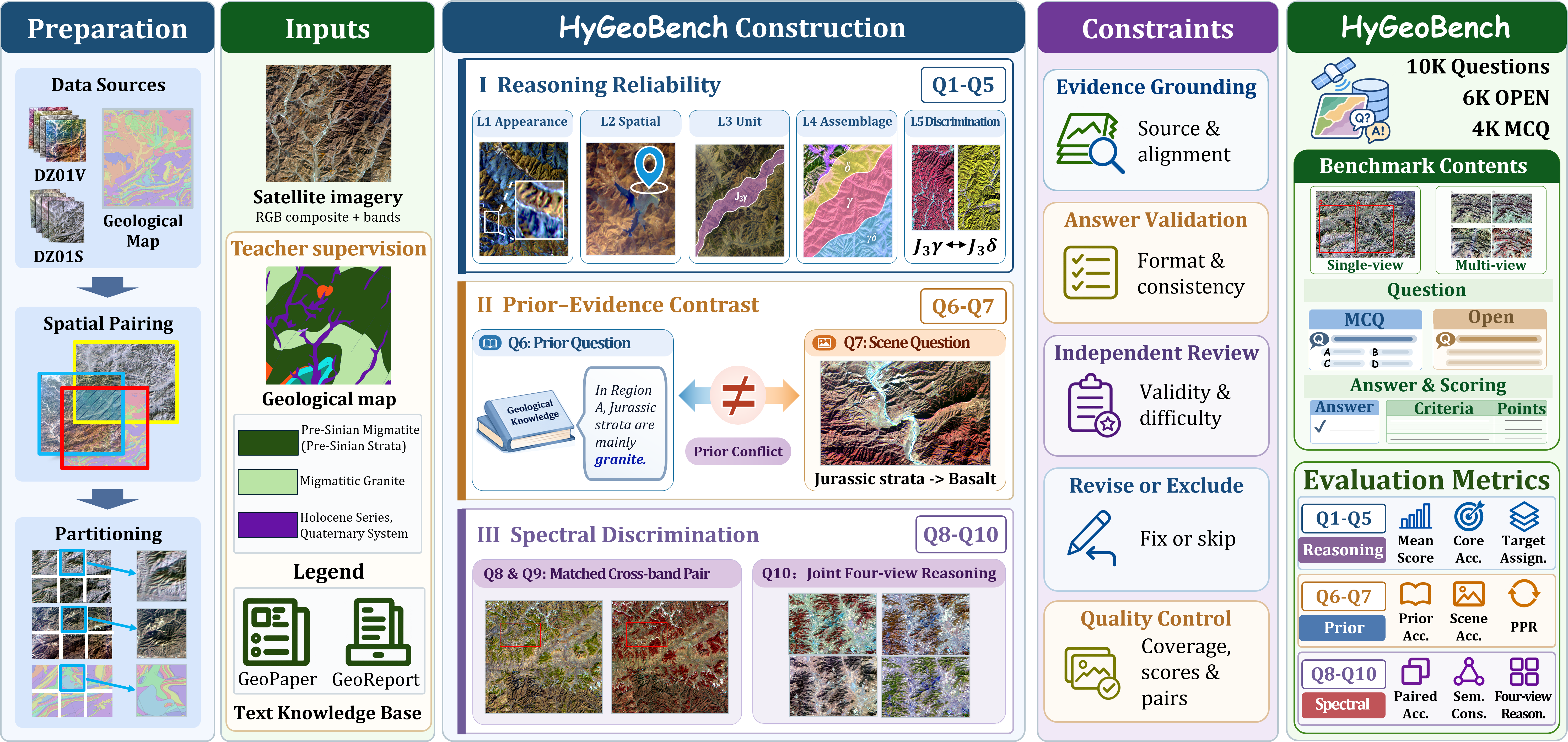}
    \caption{Overview of the \hygeo construction pipeline, illustrating data preparation, question generation, and constraint validation.}
    \label{fig:question_construction_workflow}
\end{figure*}

\section{\hygeo}
\label{sec:dataset}

We introduce \hygeo, a diagnostic benchmark for geological interpretation of RS hyperspectral imagery. It evaluates MLLMs across three dimensions: reasoning reliability, prior--evidence contrast, and spectral discrimination. As shown in Fig.~\ref{fig:question_construction_workflow}, it is built from spatially paired satellite imagery and geological maps. Geological maps, original legends, and a knowledge base provide supervision for evidence-grounded question generation, followed by independent review and quality control. The benchmark contains 10,000 questions (4,000 multiple-choice and 6,000 open-ended), covering three diagnostic categories and ten question types.

\subsection{Benchmark Overview}
\label{sec:benchmark_overview}

\hygeo is designed around hyperspectral satellite imagery to exploit its rich spectral information for lithological discrimination and geological interpretation. We compare DZ01 with EMIT~\cite{NASAEarthdataEMITL2A} and EnMAP~\cite{EnMAPMission} through lithology classification across six models and seven RGB composites under matched experimental conditions. \hygeo employs DZ01S shortwave-infrared and DZ01V visible--near-infrared satellite imagery, containing 10 and 16 spectral bands, respectively~\cite{Wang2025Geology1Mission}. DZ01 achieves the highest mean accuracy (59.92\%) and macro-F1 (48.45\%), compared with 53.7\% / 41.9\% for EMIT and 55.9\% / 43.4\% for EnMAP, respectively. Further experimental details are provided in Appendix. These results support the selection of DZ01 as the primary imagery source for \hygeo.

Geological reference information is obtained from the National 1:200,000-Scale Digital Geological Map Spatial Database~\cite{Li2019GeologicalMap,Li2019GeologicalMapDataset}. We establish spatial correspondences between satellite imagery and geological maps to construct single-sensor and dual-sensor samples with consistent geographic coverage. The former primarily supports hierarchical geological interpretation and prior--evidence contrast, whereas the latter provides corresponding views under different spectral configurations for cross-band interpretation. After validation, the DZ01S, DZ01V, and dual-sensor subsets cover 973, 1,299, and 1,456 lithological classes, respectively, with an average of 15.9, 12.6, and 11.1 classes per sample. This substantial lithological diversity highlights the challenges of fine-grained geological recognition and spatial reasoning. The selected band combinations and their selection criteria are summarized in Table~\ref{tab:band_combinations}.

\begin{table}[t]
    \centering
    \caption{Candidate RGB band combinations for single- and dual-sensor samples before manual quality screening.}
    \label{tab:band_combinations}
    \footnotesize
    \setlength{\tabcolsep}{2.5pt}
    \renewcommand{\arraystretch}{1.12}

    \begin{tabularx}{\columnwidth}{
        @{} c c c >{\raggedright\arraybackslash}X @{}
    }
        \toprule
        \textbf{R (nm)} & \textbf{G (nm)} & \textbf{B (nm)}
        & \textbf{Reference} \\
        \midrule

        \multicolumn{4}{@{}l}{\textbf{Single-sensor samples}} \\
        \midrule

        1642 & 1600 & 1196.5
        & WV-3 SWIR 3--2--1~\cite{kuester2016worldview3} \\

        1757.5 & 1600 & 1196.5
        & SWIR spectral basis~\cite{pelta2019potential,aguilar2025mapping} \\

        2147 & 1757.5 & 1600
        & HI SWIR band set~\cite{guffogg2023plastic} \\

        2213.5 & 2147 & 1600
        & ASTER 6--5--4 (SWIR)~\cite{perez2010aster} \\

        2265 & 2213.5 & 2147
        & ASTER 7--6--5 (DCS)~\cite{geoimage_swir} \\

        2335 & 2213.5 & 2147
        & ASTER 8--6--5 (SWIR)~\cite{jica2006mauritania} \\

        633.5 & 550 & 490
        & True-color RGB~\cite{nasa2013truecolor} \\

        850.5 & 633.5 & 550
        & NIR--red--green~\cite{riebeekFalseColor} \\

        850.5 & 722.5 & 633.5
        & WV-2 7--6--5~\cite{digitalglobe2010worldview2} \\

        814 & 769 & 722.5
        & Sentinel-2 7--6--5~\cite{michel2022sen2venus,copernicusSentinel2Mission} \\

        1000.5 & 850.5 & 722.5
        & Geological NIR analogue~\cite{allender2020exomars} \\

        850.5 & 814 & 769
        & OIF-selected bands~\cite{chavez1984oif,alnahmi2017geological} \\

        \midrule
        \multicolumn{4}{@{}l}{\textbf{Dual-sensor samples}} \\
        \midrule

        633.5 & 550 & 490
        & True-color RGB~\cite{nasa2013truecolor} \\

        850.5 & 678.5 & 550
        & NIR--red--green~\cite{riebeekFalseColor} \\

        1642 & 2213.5 & 2335
        & ASTER 4--6--8 (SWIR)~\cite{abrams2002aster,nouman2024aster} \\

        2335 & 2265 & 1642
        & ASTER 8--7--4 (SWIR)~\cite{rajendran2015moho} \\

        1600 & 850.5 & 678.5
        & Sentinel-2 11--8--4~\cite{copernicusSentinel2Mission} \\

        2213.5 & 850.5 & 678.5
        & Landsat 9 7--5--4~\cite{usgsLandsat9} \\

        2213.5 & 1600 & 490
        & SWIR2--SWIR1--blue~\cite{kamel2016geologic,seleim2022lithological} \\

        \bottomrule
    \end{tabularx}
\end{table}

Based on these data, we design three complementary diagnostic task categories. The first, \textbf{Reasoning Reliability}, assesses five capability levels from visual observation to geological judgment (Q1--Q5). Visual Observation examines the description of target tone, texture, and morphology. Spatial Topology evaluates adjacency, connectivity, containment, and separation among targets. Geological Unit Identification assesses the mapping of image regions to geological units. Lithological Assemblage Analysis examines the identification of major lithological constituents within target units. Fine-grained Discrimination evaluates the distinction between similar units using appearance, compositional, and spatial cues. Comparing responses within the same scene reveals how stage-wise performance relates to final geological judgments.

The second category, \textbf{Prior--Evidence Contrast}, examines whether models make scene-specific judgments when geological priors conflict with local evidence~\cite{guan2024hallusionbench}. Q6 and Q7 correspond to General Geological Prior and Scene-Specific Questions, respectively. They address comparable geological properties but have different reference conclusions: Q6 tests general knowledge without images, whereas Q7 requires judgments from target appearance, distribution, and spatial relationships. This pairing assesses whether models ground prior knowledge in local evidence rather than follow conflicting general expectations.

The third category, \textbf{Spectral Discrimination}, evaluates how models handle band-dependent appearance changes~\cite{cudahy2016mineralmapping}. Q8--Q9 form matched cross-band pairs with fixed targets, questions, and reference answers but different band combinations. Q10 provides four band composites of the same region for joint geological unit identification and spatial reasoning. Together, these tasks assess whether models integrate complementary spectral views without mistaking imaging variations for geological differences.

These complementary tasks probe capability differences across interpretation levels, the grounding of prior knowledge in local evidence, and the reliability of judgments across spectral configurations. \hygeo therefore supports diagnostic analysis beyond final-answer accuracy.

\subsection{Benchmark Construction}
\label{sec:question_set_construction}

As shown in Fig.~\ref{fig:question_construction_workflow}, the construction workflow comprises data preparation, question generation, and constraint validation. We collect and verify DZ01S/DZ01V imagery and geological maps, harmonize their coordinate reference systems, and identify their common valid coverage. Single-sensor samples use the overlap between the selected imagery and geological map, while dual-sensor samples use the common coverage of both sensors and the geological map.

The imagery and geological labels are partitioned using identical geographic windows, followed by the generation of RGB composites with different band combinations. Single-sensor samples retain native spatial grids, while temporally matched dual-sensor observations cover the same geological map regions. Automated checks and manual screening verify image quality and image--label correspondence.

During question generation, satellite imagery serves as visual input, while geological maps, original legends, and a knowledge base provide supervision for question generation. The knowledge base contains 85 geological reports and 120 research papers. These materials are used only for question construction and reference-answer verification and are not provided to evaluated models. Questions are designed according to visible geological targets, spatial relationships, and verified geological knowledge, covering reasoning reliability, prior--evidence contrast, and spectral discrimination.

For prior--evidence pairs, the two questions must address comparable geological judgments but have different evidence-supported reference conclusions. Their answers are normalized into a shared semantic space for comparison. Pair validity requires three criteria: semantic comparability, verified support for the general geological prior, and sufficient geological annotations and image evidence for the scene-specific conclusion. Only pairs satisfying all criteria with different normalized reference conclusions are retained.

Questions are generated according to task-specific evidence and discrimination constraints. Given an evidence set $\mathcal{E}_i$, task objective $t_i$, response format $f_i$, and constraints $\mathcal{C}$, the generation process is defined as
\begin{equation}
    (q_i,y_i,\mathcal{J}_i,\mathcal{R}_i)
    \sim G_{\theta}(\mathcal{E}_i,t_i,f_i;\mathcal{C}),
    \label{eq:adaptive_question_generation}
\end{equation}
where $q_i$ denotes the question, $y_i$ its reference answer, $\mathcal{J}_i$ the normalized judgment set, and $\mathcal{R}_i$ the scoring rules. To prevent answer shortcuts, image-based multiple-choice questions combine geological identification with local visual or spatial relationships, requiring both judgments for a correct answer. Open-ended reference answers are decomposed into independently scorable components to support partial credit. All responses use a ten-point scale: multiple-choice answers receive either 10 or 0 points, while open-ended scores sum the points awarded to individual components, whose maximum weights sum to 10. Scoring criteria cover target localization, geological identity, lithological composition, spatial relationships, and supporting explanations.

\subsection{Validation and Diagnostic Evaluation}
\label{sec:benchmark_scoring_quality}

\begin{figure*}[t]
    \centering
    \includegraphics[width=0.95\textwidth]{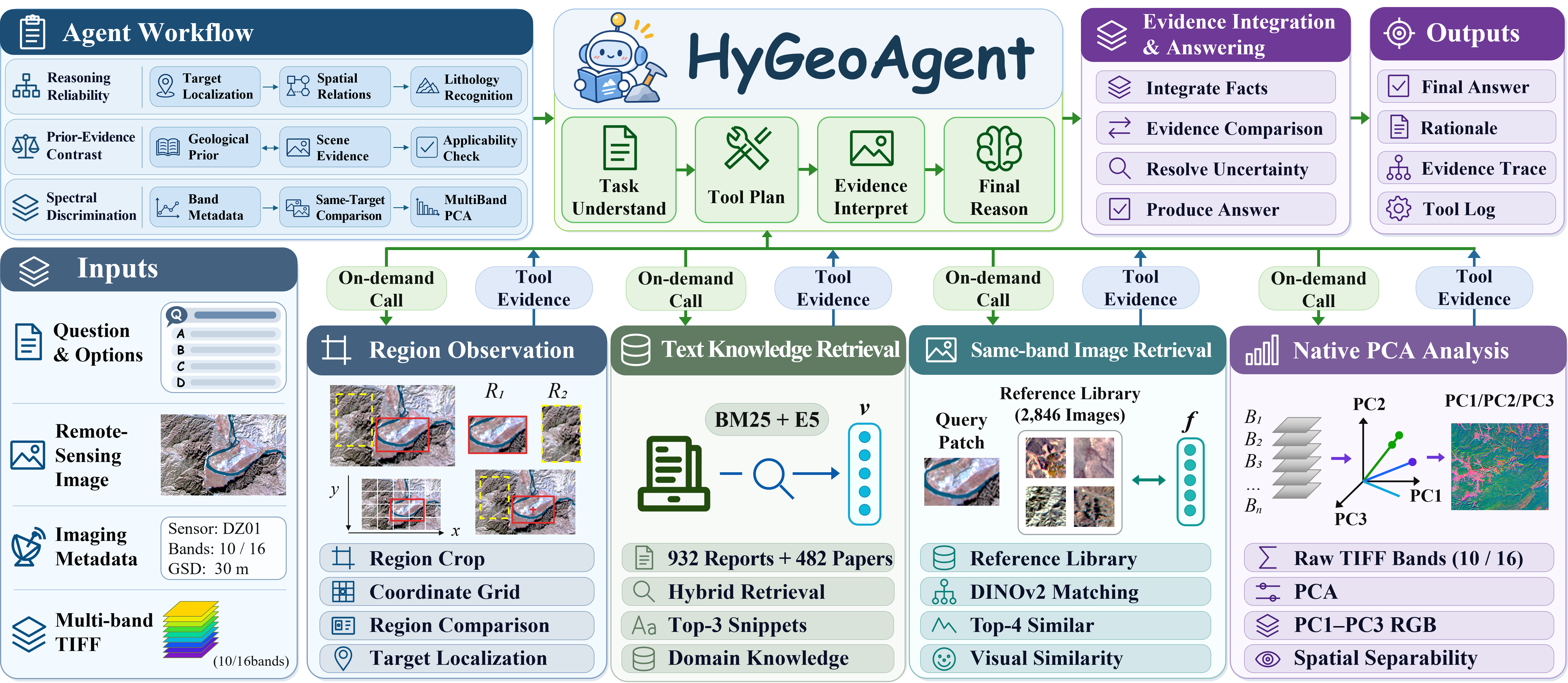}
    \caption{Overview of \hygeoagent, illustrating adaptive tool use for region observation, text knowledge retrieval, same-band image retrieval, and native PCA analysis.}
    \label{fig:hygeoagent_architecture}
\end{figure*}

\hygeo employs automated validation, independent model-based review, and expert assessment for quality control. Review criteria cover source reliability, question--answer consistency, task clarity, and paired-question validity. Question quality is assessed through correctness, answerability, scoring validity, scientific applicability, and diversity, while design difficulty covers fine-grained discrimination, spatial reasoning, evidence integration, and resistance to shortcut answers. Expert screening ensures balanced coverage of ten question types and diverse input images, reducing 14,488 initial questions to 10,000 while improving question quality and design difficulty. The final benchmark contains 5,614 unique input images. Further details on quality improvement are provided in Appendix.

Beyond conventional answer scoring, we evaluate three aspects of geological interpretation corresponding to RQ1--RQ3. For \textbf{Reasoning Reliability} (RQ1), Q1--Q5 assess five stages of geological interpretation: visual observation, spatial localization, geological map-unit identification, lithological assemblage interpretation, and fine-grained classification. To capture how earlier-stage weaknesses affect subsequent judgments, we adopt a cumulative minimum-based evaluation. Let $\bar{s}_{ij}$ denote the mean percentage score assigned by the judges to case $i$ at stage $j$. Inspired by group-wise evaluation~\cite{guan2024hallusionbench} and minimum-based fuzzy intersection~\cite{zadeh1965fuzzysets}, we define
\begin{equation}
    S_{\mathrm{RQ1}}
    =\frac{1}{5N}\sum_{i=1}^{N}\sum_{k=1}^{5}
      \min_{1\leq j\leq k}\bar{s}_{ij}.
    \label{eq:rq1_cumulative}
\end{equation}
Here, $N$ denotes the number of valid cases with complete responses and valid scores across all five stages. For each case, the cumulative minimum captures the weakest performance across the first $k$ stages. Averaging these values reflects stage-wise weaknesses and their potential impact on subsequent judgments, thereby assessing the reliability of the overall interpretation process.

For \textbf{Prior--Evidence Contrast} (RQ2), we compare normalized predictions on Q6 and Q7. Since their reference conclusions differ, we measure whether models retain correctly identified priors when answering Q7. The conditional Prior Persistence Rate (PPR) is calculated over valid pairs $V$:
\begin{equation}
    \mathrm{PPR}=
    \frac{
    \sum_{i\in V}
    \mathbf{1}[\hat{k}_i=k_i^*]
    \mathbf{1}[\hat{l}_i=k_i^*]
    }{
    \sum_{i\in V}
    \mathbf{1}[\hat{k}_i=k_i^*]
    },
    \label{eq:prior_persistence}
\end{equation}
where $k_i^{*}$ and $l_i^{*}$ are the reference conclusions for general-prior and scene-specific questions, respectively, and $\hat{k}_i$ and $\hat{l}_i$ are the corresponding predictions. The denominator selects pairs with correctly identified priors, while the numerator counts cases where the prior persists in Q7. Higher PPR indicates stronger prior persistence and weaker adaptation to conflicting visual evidence. For clearer presentation, we report RQ2 as $100(1-\mathrm{PPR})$, where higher scores indicate better adaptation to conflicting visual evidence.

For \textbf{Spectral Discrimination} (RQ3), Q8 and Q9 evaluate corresponding targets under different band composites, while Q10 involves four-view interpretation. These questions assess cross-band recognition and joint reasoning over complementary spectral views. Let $\bar{s}_{iq}$ denote the mean judge score for question $q$ in group $i$, and $|G|=622$ the number of common Q8--Q10 groups. Following group-wise evaluation~\cite{guan2024hallusionbench}, we adopt the algebraic product of fuzzy sets~\cite{zadeh1965fuzzysets} to measure joint performance:
\begin{equation}
    S_{\mathrm{RQ3}}
    =\frac{100}{|G|}\sum_{i\in G}
      \prod_{q=8}^{10}\frac{\bar{s}_{iq}}{100}.
    \label{eq:rq3_joint_score}
\end{equation}
The product penalizes weak performance on any constituent question, emphasizing consistent cross-band interpretation. For presentation, we multiply RQ3 scores by 10, with higher values indicating better joint performance.
% HyGeoAgent -- revised English main text for CVPR.
% Structure follows the revised Chinese manuscript:
%   4.1 Agent Workflow; 4.2 Geological Tools.
% The original framework figure and BOTH original equation environments
% (including their labels) are reproduced unchanged.
% The original Section 4 contains NO table environments to move.
% Required packages are those already used by the original manuscript
% (e.g., graphicx and amsmath). Uses existing \hygeoagent macro and BibTeX keys.

\section{\hygeoagent}
\label{sec:agent}

\subsection{Agent Workflow}
\label{sec:agent_framework}

To address challenges in geological reasoning, knowledge grounding, and band-dependent interpretation, we propose \hygeoagent, a tool-augmented framework for geological interpretation. As shown in Fig.~\ref{fig:hygeoagent_architecture}, a single MLLM coordinates task understanding, adaptive tool selection, and answer generation through iterative interactions~\cite{yao2023react}. It receives questions, options, remote sensing imagery, and metadata, with native multiband data available when needed. These inputs help identify geological targets, determine interpretation requirements, and assess imaging conditions. Problem-oriented constraints guide tool use, while acquired observations and unresolved questions inform subsequent actions, progressively refining geological judgments.

For \textbf{Reasoning Reliability}, the model uses coordinate grids and region cropping to localize targets and inspect visual features, followed by paired-region comparison to assess spatial relationships. For geological-unit identification, same-band image retrieval provides labeled visual references, while text knowledge retrieval~\cite{lewis2020rag} supplies candidate properties and lithological constituents. Fine-grained discrimination further compares visual features, lithological combinations, and spatial distributions to identify specific geological categories.

For \textbf{Prior--Evidence Contrast}, the model distinguishes general-knowledge questions from scene-specific judgments. Text knowledge retrieval verifies geological concepts, typical characteristics, and applicability conditions, while region observation and same-band image retrieval provide local evidence and comparable references. When geological priors conflict with visual observations, the model assesses their applicability against scene-specific evidence rather than directly applying general expectations.

For \textbf{Spectral Discrimination}, the model uses RGB center wavelengths, spatial resolution, and spatial correspondence to identify and compare targets across band composites. Same-band image retrieval provides matched references, while text knowledge retrieval supports candidate assessment and interpretation of band-dependent variations. When target boundaries or internal structures remain ambiguous, native PCA analysis generates complementary representations from native spectral bands.

During execution, the MLLM sends tool requests to an execution program, which returns text, region crops, reference images, or PCA composites as multimodal inputs. The model compares these outputs with previous observations and candidate interpretations to determine whether further inspection is needed. Uncertain target locations trigger region observation, ambiguous candidates prompt text knowledge or image retrieval, and unresolved spectral patterns may invoke native PCA analysis. Each question allows up to six interaction rounds, with multiple calls per round. The model then produces a geological interpretation and supporting explanation based on the acquired visual observations and geological knowledge.

\subsection{Geological Tools}
\label{sec:agent_tools}

\begin{table*}[!t]
    \centering
    \caption{
        Benchmark results for nineteen base models, three
        \hygeoagent configurations, GeoMM-Agent, and a text-only baseline.
    }
    \label{tab:overall_benchmark}

    \begingroup
    \scriptsize
    \setlength{\tabcolsep}{2pt}
    \renewcommand{\arraystretch}{1.12}

    \resizebox{\textwidth}{!}{%
    \begin{tabular}{@{}l*{26}{r}@{}}
        \toprule
        \multirow{2}{*}{\textbf{Model}}
        & \multicolumn{2}{c}{\textbf{Q1}}
        & \multicolumn{2}{c}{\textbf{Q2}}
        & \multicolumn{2}{c}{\textbf{Q3}}
        & \multicolumn{2}{c}{\textbf{Q4}}
        & \multicolumn{2}{c}{\textbf{Q5}}
        & \multicolumn{2}{c}{\textbf{Q6}}
        & \multicolumn{2}{c}{\textbf{Q7}}
        & \multicolumn{2}{c}{\textbf{Q8}}
        & \multicolumn{2}{c}{\textbf{Q9}}
        & \multicolumn{2}{c}{\textbf{Q10}}
        & \multirow{2}{*}{\textbf{MCQ}}
        & \multirow{2}{*}{\textbf{Open}}
        & \multirow{2}{*}{\textbf{RQ1}}
        & \multirow{2}{*}{\textbf{RQ2}}
        & \multirow{2}{*}{\textbf{RQ3}}
        & \multirow{2}{*}{\textbf{OA}} \\

        & MCQ & Open
        & MCQ & Open
        & MCQ & Open
        & MCQ & Open
        & MCQ & Open
        & MCQ & Open
        & MCQ & Open
        & MCQ & Open
        & MCQ & Open
        & MCQ & Open
        & & & & & & \\
        \midrule

        % ================================
        % LLMs: Only Text as Input
        % ================================
        \multicolumn{27}{c}{\textbf{\textit{Only Text as Input}}} \\
        \midrule

        DeepSeek V4.1 Flash~\cite{deepseek2026v41flash}
        & 54.60 & 26.37
        & 44.75 & 17.18
        & 48.32 & 3.35
        & 57.56 & 3.23
        & 37.82 & 7.37
        & 91.63 & 87.09
        & 48.80 & 4.35
        & 45.80 & 4.15
        & 45.42 & 3.67
        & 41.86 & 6.03
        & 52.30 & 17.31 & 9.19 & 22.61 & 37.15 & 31.31 \\

        \midrule

        % ================================
        % Open-source MLLMs
        % ================================
        \multicolumn{27}{c}{\textbf{\textit{Open-source MLLMs}}} \\
        \midrule

        EarthDial-4B-RGB~\cite{earthdial2025}
        & 28.10 & 8.58
        & 26.82 & 8.34
        & 28.85 & 1.66
        & 33.61 & 0.89
        & 26.96 & 3.18
        & 59.09 & 18.84
        & 32.38 & 1.37
        & 25.06 & 1.65
        & 24.81 & 1.66
        & 23.13 & 1.29
        & 31.51 & 5.11 & 4.24 & 22.12 & 11.56 & 15.67 \\

        SkyEyeGPT~\cite{skyeyegpt2025}
        & 25.24 & 1.26
        & 20.52 & 0.92
        & 25.98 & 1.27
        & 28.36 & 0.76
        & 29.20 & 1.53
        & 39.71 & 16.78
        & 23.60 & 0.78
        & 21.50 & 1.19
        & 20.48 & 0.85
        & 23.26 & 0.89
        & 26.27 & 2.71 & 2.33 & 14.74 & 6.50 & 12.13 \\

        TEOChat-7B~\cite{teochat2025}
        & 30.20 & 18.17
        & 22.48 & 14.39
        & 25.00 & 3.37
        & 30.81 & 2.74
        & 29.20 & 8.01
        & 58.45 & 29.65
        & 26.71 & 3.07
        & 23.03 & 4.86
        & 23.54 & 4.54
        & 24.68 & 2.39
        & 29.94 & 9.75 & 5.68 & 15.92 & 14.22 & 17.83 \\

        GeoChat-7B~\cite{geochat2024}
        & 28.38 & 20.93
        & 21.64 & 17.74
        & 27.03 & 4.05
        & 32.00 & 2.60
        & 28.36 & 7.90
        & 53.51 & 32.92
        & 24.24 & 3.47
        & 23.66 & 5.48
        & 23.28 & 4.95
        & 20.93 & 2.40
        & 28.92 & 10.98 & 5.68 & 15.69 & 7.78 & 18.16 \\

        LHRS-Bot-Nova~\cite{lhrsbotnova2025}
        & 31.03 & 25.01
        & 26.05 & 27.53
        & 33.89 & 7.71
        & 39.29 & 5.87
        & 31.51 & 15.17
        & 74.64 & 47.37
        & 29.35 & 6.78
        & 29.39 & 9.59
        & 30.79 & 8.99
        & 27.13 & 6.16
        & 35.83 & 16.96 & 7.74 & 18.35 & 17.94 & 24.51 \\

        \midrule

        Molmo2-8B~\cite{clark2026molmo2}
        & 47.49 & 37.39
        & 34.45 & 46.87
        & 40.13 & 11.44
        & 48.95 & 8.51
        & 35.29 & 15.19
        & 88.52 & 59.14
        & 39.23 & 8.18
        & 37.02 & 11.64
        & 35.50 & 11.03
        & 36.56 & 9.45
        & 45.03 & 23.36 & 12.95 & 24.09 & 39.56 & 32.03 \\

        InternVL3.5-8B~\cite{wang2025internvl35}
        & 42.33 & 31.21
        & 34.73 & 38.62
        & 34.80 & 6.88
        & 46.85 & 6.41
        & 34.03 & 10.66
        & 89.23 & 52.62
        & 37.56 & 5.40
        & 32.06 & 7.30
        & 31.68 & 7.41
        & 35.66 & 5.72
        & 42.68 & 18.57 & 10.71 & 22.72 & 26.98 & 28.22 \\

        LLaVA-OneVision-2~\cite{an2026llavaov2}
        & 42.05 & 26.33
        & 34.87 & 35.75
        & 37.25 & 10.10
        & 46.85 & 6.70
        & 32.98 & 13.24
        & 87.80 & 59.96
        & 39.63 & 7.27
        & 37.66 & 10.28
        & 39.82 & 9.59
        & 37.86 & 7.24
        & 43.94 & 19.77 & 9.56 & 24.46 & 38.33 & 29.44 \\

        Qwen3.5-9B~\cite{qwen2026qwen35}
        & 53.28 & 42.91
        & 38.45 & 49.97
        & 35.50 & 9.83
        & 48.53 & 8.87
        & 31.72 & 13.23
        & 95.61 & 67.17
        & 40.83 & 7.08
        & 35.50 & 9.97
        & 36.26 & 9.46
        & 32.56 & 8.10
        & 45.77 & 24.43 & 14.32 & 24.11 & 24.39 & 32.97 \\

        \midrule

        % ================================
        % Closed-source MLLMs
        % ================================
        \multicolumn{27}{c}{\textbf{\textit{Closed-source MLLMs}}} \\
        \midrule

        Claude Opus 5.5~\cite{anthropic2026opus55}
        & 73.43 & 73.56
        & 74.58 & 81.00
        & 54.41 & 13.54
        & \textbf{67.86} & 13.19
        & 45.59 & 15.90
        & 99.28 & 92.67
        & \textbf{55.26} & 10.92
        & \textbf{48.09} & 21.03
        & 46.95 & 21.27
        & 46.77 & 18.92
        & 63.02 & 38.32 & 31.63 & 29.59 & 64.40 & 48.20 \\

        Gemini 3.8~\cite{google2026gemini38}
        & 72.80 & 69.43
        & 71.15 & 78.77
        & 55.67 & 17.46
        & 64.71 & 15.94
        & 44.54 & 23.52
        & \textbf{99.76} & \textbf{92.89}
        & 49.92 & 14.56
        & 36.26 & 22.02
        & 34.73 & \textbf{21.94}
        & 48.45 & \textbf{21.50}
        & 60.21 & 39.94 & 30.57 & 29.30 & 65.33 & 48.05 \\

        Grok 4.7~\cite{spacexsi2026grok47}
        & 69.04 & 64.93
        & 60.50 & 72.85
        & 47.06 & 10.39
        & 56.93 & 9.90
        & 36.34 & 6.43
        & 99.52 & 89.01
        & 50.72 & 7.77
        & 43.51 & 13.20
        & 38.55 & 13.47
        & \textbf{48.84} & 14.69
        & 56.38 & 32.42 & 25.40 & 26.10 & 51.34 & 42.00 \\

        Qwen3.8 Max~\cite{qwen2026qwen38max}
        & 72.18 & \textbf{76.36}
        & 68.07 & 82.65
        & 52.73 & 17.72
        & 64.29 & 16.64
        & 41.60 & 23.54
        & 99.28 & 89.48
        & 48.09 & 13.47
        & 30.15 & 21.70
        & 32.44 & 21.51
        & 44.57 & 20.40
        & 57.98 & 40.75 & 30.08 & 28.38 & 48.09 & 47.64 \\

        DeepSeek V4.1 Flash~\cite{deepseek2026v41flash}
        & 52.72 & 64.04
        & 38.66 & 68.20
        & 41.32 & 13.35
        & 44.05 & 12.20
        & 34.45 & 15.86
        & 91.63 & 87.09
        & 38.60 & 10.16
        & 30.92 & 15.87
        & 30.41 & 15.04
        & 33.72 & 14.88
        & 44.96 & 33.84 & 20.04 & 22.38 & 23.31 & 38.29 \\

        Kimi K3.1 Pro~\cite{moonshot2026kimik31pro}
        & 70.50 & 74.05
        & 60.08 & 79.55
        & 49.37 & 16.21
        & 60.71 & 14.37
        & 39.71 & 20.86
        & 99.52 & 90.23
        & 47.13 & 11.43
        & 35.11 & 19.61
        & 38.93 & 19.71
        & 40.31 & 18.97
        & 56.17 & 38.86 & 28.63 & 26.71 & 43.21 & 45.79 \\

        GLM 5.3 Flash~\cite{zai2026glm53flash}
        & 60.46 & 68.53
        & 46.43 & 74.69
        & 47.41 & 14.14
        & 55.25 & 12.77
        & 32.49 & 18.74
        & 96.65 & 91.63
        & 40.19 & 11.52
        & 25.57 & 17.02
        & 32.82 & 17.35
        & 36.43 & 16.54
        & 49.31 & 36.61 & 23.82 & 23.49 & 32.76 & 41.69 \\

        Doubao Seed 2.1 Pro~\cite{bytedance2026seed21}
        & 69.67 & 74.38
        & 66.39 & 81.83
        & 45.17 & 13.43
        & 58.82 & 12.18
        & 41.04 & 16.59
        & 99.52 & 89.14
        & 47.37 & 9.39
        & 41.09 & 17.28
        & 39.69 & 17.69
        & 43.41 & 15.58
        & 56.92 & 37.24 & 27.67 & 25.82 & 44.74 & 45.12 \\

        Nemotron 3.5~\cite{nvidia2026nemotron35}
        & 54.81 & 62.19
        & 49.79 & 71.68
        & 39.71 & 14.75
        & 58.61 & 12.84
        & 40.34 & 13.38
        & 98.56 & 90.66
        & 47.13 & 10.32
        & 35.88 & 16.65
        & 39.69 & 17.08
        & 37.60 & 15.92
        & 51.58 & 34.62 & 21.53 & 25.87 & 36.46 & 41.40 \\

        GPT-6.1 Sol~\cite{openai2026gpt61sol}
        & 66.39 & 69.27
        & 56.93 & 75.37
        & 46.92 & 10.72
        & 56.72 & 10.10
        & 40.06 & 10.96
        & 99.04 & 88.13
        & 47.69 & 7.52
        & 40.46 & 14.82
        & 43.13 & 13.62
        & 41.47 & 11.25
        & 55.29 & 33.61 & 24.96 & 24.65 & 42.10 & 42.28 \\

        \midrule

        \multicolumn{27}{c}{\textbf{\textit{Agent}}} \\
        \midrule

        GeoMM-Agent~\cite{geommbench2026}
        & 55.23 & 66.57
        & 45.17 & 68.26
        & 40.13 & 12.28
        & 52.52 & 11.37
        & 37.61 & 13.56
        & 99.28 & 86.48
        & 42.34 & 10.56
        & 32.44 & 15.67
        & 38.17 & 14.37
        & 31.78 & 12.68
        & 48.95 & 33.43 & 21.44 & 24.52 & 25.12 & 39.64 \\

        \textbf{\hygeoagent DeepSeek}
        & 61.68 & 75.45
        & 59.07 & 80.89
        & 44.63 & 17.04
        & 59.79 & 16.25
        & 39.16 & 20.25
        & \textbf{99.76} & 92.86
        & 46.51 & 13.90
        & 42.64 & 21.91
        & 38.04 & 19.94
        & 46.88 & 21.23
        & 55.10 & 40.32 & 28.26 & 27.60 & 52.92 & 46.23 \\

        \textbf{\hygeoagent GPT}
        & 70.71 & 75.09
        & 68.49 & 81.99
        & 45.31 & 13.02
        & 56.72 & 13.99
        & 36.48 & 13.25
        & \textbf{99.76} & 88.43
        & 44.98 & 10.01
        & 38.17 & 19.30
        & 39.57 & 15.63
        & 44.19 & 16.93
        & 56.15 & 37.19 & 29.15 & 25.21 & 44.43 & 44.77 \\

        \textbf{\hygeoagent Claude}
        & \textbf{76.78} & 75.72
        & \textbf{75.84} & \textbf{83.61}
        & \textbf{57.77} & \textbf{17.87}
        & 66.11 & \textbf{17.98}
        & \textbf{48.35} & \textbf{24.33}
        & 99.52 & 91.75
        & 51.83 & \textbf{16.92}
        & 45.80 & \textbf{23.44}
        & \textbf{48.09} & 20.94
        & 48.71 & 21.38
        & \textbf{63.81} & \textbf{41.82}
        & \textbf{34.63} & \textbf{31.76}
        & \textbf{71.66} & \textbf{50.62} \\

        \bottomrule
    \end{tabular}%
    }
    \endgroup
\end{table*}

To support the workflow, \hygeoagent integrates four tools for visual inspection, geological knowledge retrieval, reference image comparison, and spectral analysis.

\noindent\textbf{Region Observation.}\label{sec:agent_region_observation}
This tool uses coordinate grids, region cropping, and paired-region comparison to inspect target features while preserving scene context. Crops reveal brightness, texture, morphology, and boundaries, while paired views assess spatial continuity, adjacency, and orientation. Target crops can also be passed to same-band image retrieval.

\noindent\textbf{Text Knowledge Retrieval.}\label{sec:agent_text_retrieval}
We construct a knowledge base of 1,429 geological reports, research papers, and reference documents, independent of the sources used to build \hygeo. The tool combines lexical~\cite{sqliteFts5} and semantic retrieval~\cite{wang2024multilinguale5}, followed by reciprocal rank fusion~\cite{cormack2009rrf} and geological entity- and concept-based reranking. Retrieved passages cover geological definitions, lithological constituents, unit characteristics, and distinguishing properties, helping verify geological knowledge and its applicability to target scenes.

\noindent\textbf{Same-Band Image Retrieval.}\label{sec:agent_image_retrieval}
The reference library contains 2,846 satellite images from 642 geographic samples across 321 geological-label categories. Given a target crop, the tool filters references by identical ordered RGB center wavelengths and ranks them using visual features and cosine similarity. It retrieves up to four top-ranked images and four additional references prioritizing label diversity. Geological categories among the 30 nearest neighbors are also summarized as supplementary information. These references support visual comparison and geological candidate discrimination under matched imaging conditions.

\noindent\textbf{Native PCA Analysis.}\label{sec:agent_pca}
Using native single-band TIFFs, the tool applies mean-centered covariance PCA~\cite{richards2022remote} separately to DZ01S (10 bands) and DZ01V (16 bands):
\begin{equation}
    \mathbf{Z}
    =(\mathbf{X}-\mathbf{1}\boldsymbol{\mu}^{\top})
      \mathbf{U}_{1:3},
    \label{eq:agent_native_pca}
\end{equation}
where $\mathbf{X}$ contains native spectral vectors, $\boldsymbol{\mu}$ is the fitted mean, and $\mathbf{U}_{1:3}$ contains the first three covariance eigenvectors. The resulting RGB composite highlights spectral variations, target boundaries, and local structures, providing complementary geological information~\cite{mohamed2021geologicalPCA}.

Together, these tools provide complementary visual, geological, and spectral information. Through adaptive selection and iterative comparison, \hygeoagent supports hierarchical interpretation, geological knowledge verification, and cross-band analysis.

\begin{table}[!t]
    \centering
    \caption{Ablation results for different agent configurations. Red arrows show the decrease relative to the Full Agent in each score column.}
    \label{tab:ablation_study}

    \small
    \renewcommand{\arraystretch}{1.12}
    \setlength{\tabcolsep}{4pt}

    \resizebox{0.95\columnwidth}{!}{%
    \begin{tabular}{lccc}
        \toprule
        \textbf{Configuration} & \textbf{MCQ} & \textbf{Open}
        & \textbf{Average Score} \\
        \midrule
        \textbf{Full Agent}
        & \textbf{62.50} & \textbf{43.42} & \textbf{51.05} \\
        \midrule
        Original Model
        & 55.00\,{\color{red}$\downarrow 7.50$}
        & 39.75\,{\color{red}$\downarrow 3.67$}
        & 45.85\,{\color{red}$\downarrow 5.20$} \\

        w/o Text Retrieval
        & 52.50\,{\color{red}$\downarrow 10.00$}
        & 43.04\,{\color{red}$\downarrow 0.38$}
        & 46.83\,{\color{red}$\downarrow 4.22$} \\

        w/o Reference Image Retrieval
        & 57.50\,{\color{red}$\downarrow 5.00$}
        & 41.38\,{\color{red}$\downarrow 2.04$}
        & 47.83\,{\color{red}$\downarrow 3.22$} \\

        w/o PCA Tool
        & 60.00\,{\color{red}$\downarrow 2.50$}
        & 40.29\,{\color{red}$\downarrow 3.13$}
        & 48.18\,{\color{red}$\downarrow 2.87$} \\

        w/o Local Image Observation
        & 55.00\,{\color{red}$\downarrow 7.50$}
        & 40.75\,{\color{red}$\downarrow 2.67$}
        & 46.45\,{\color{red}$\downarrow 4.60$} \\

        Static PCA Images
        & 57.50\,{\color{red}$\downarrow 5.00$}
        & 41.00\,{\color{red}$\downarrow 2.42$}
        & 47.60\,{\color{red}$\downarrow 3.45$} \\

        Static Band-wise Grayscale Images
        & 55.00\,{\color{red}$\downarrow 7.50$}
        & 42.38\,{\color{red}$\downarrow 1.04$}
        & 47.43\,{\color{red}$\downarrow 3.62$} \\
        \bottomrule
    \end{tabular}%
    }
\end{table}

\section{Experiments}
\label{sec:experiments}

\begin{figure*}[!t]
    \centering
    \includegraphics[width=\textwidth]{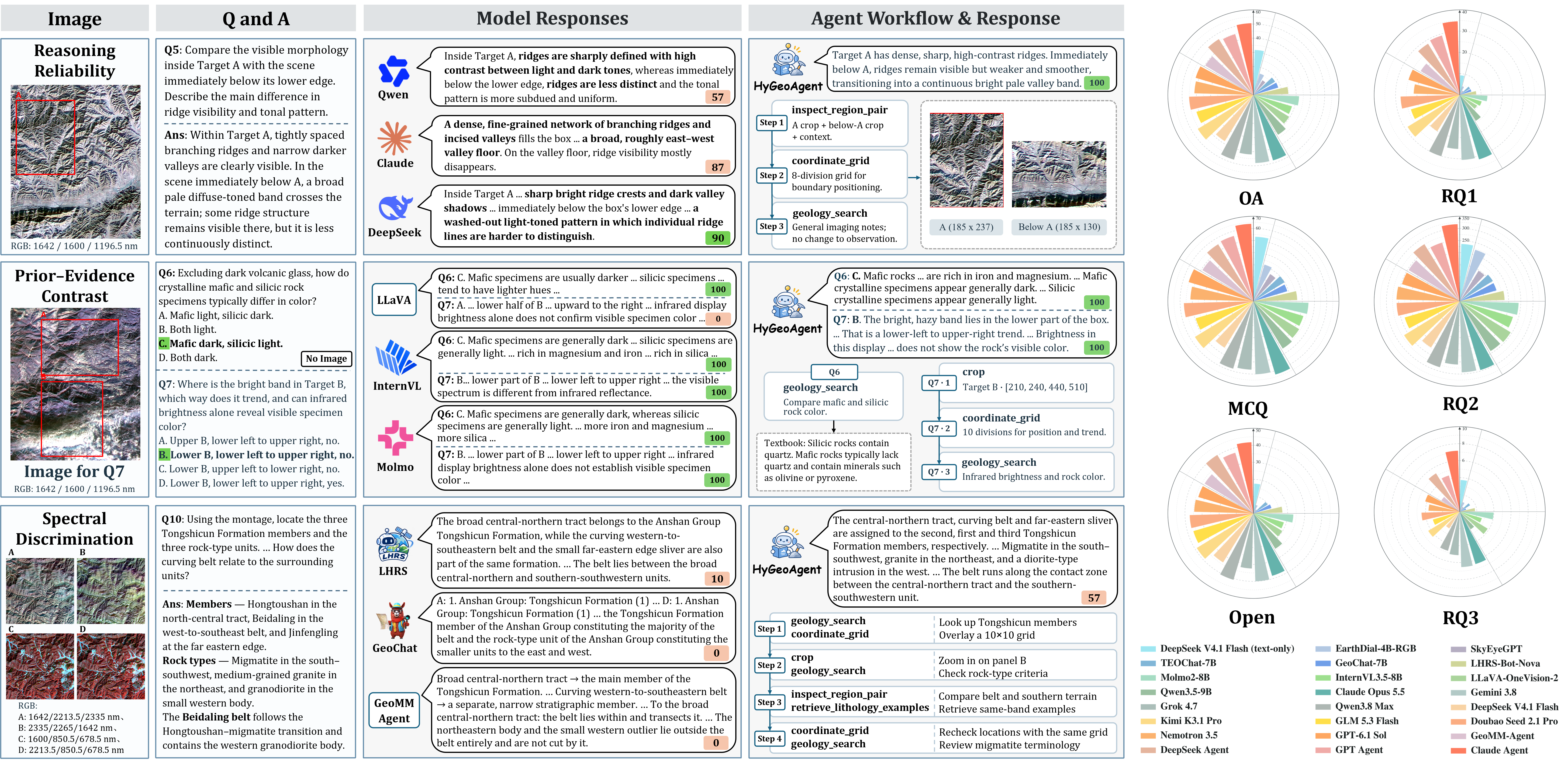}
    \caption{Qualitative comparison of model responses and \hygeoagent workflows across three diagnostic dimensions, with corresponding performance distributions.}
    \label{fig:qualitative_comparison}
\end{figure*}

\subsection{Setup and Evaluation}
\label{sec:experimental_setup}

We evaluate 20 MLLMs and agents, including remote-sensing and general-purpose open- and closed-source models. Locally deployed models with fewer than 10B parameters run on NVIDIA A800 GPUs, while others are accessed via APIs. We additionally evaluate three \hygeoagent configurations based on Claude Opus 5.5~\cite{anthropic2026opus55}, DeepSeek V4.1 Flash~\cite{deepseek2026v41flash}, and GPT-6.1 Sol~\cite{openai2026gpt61sol}, each compared with its corresponding direct-answer configuration. Further experimental details are provided in Appendix.

We use GPT-6.1 Sol, DeepSeek V4.1 Flash, and Kimi K3.1 Pro~\cite{moonshot2026kimik31pro} as three judges with shared reference answers and atomic scoring rubrics. Each response receives the mean score of the three judges. Across all responses, the mean inter-judge score standard deviation is 1.64 points, with 0.12 for MCQ and 2.65 for Open. The mean absolute score differences among the three judge pairs range from 2.19 to 2.61 points, indicating close overall agreement.

Following the evaluation protocols in Section~\ref{sec:benchmark_scoring_quality}, we report RQ1 and RQ3 using their respective diagnostic scores. Table~\ref{tab:overall_benchmark} reports Q1--Q10 scores separately for MCQ and Open, their aggregate means, and the overall accuracy (OA), calculated as the mean score across Q1--Q10.

\subsection{Results and Analysis}
\label{sec:results_analysis}

Table~\ref{tab:overall_benchmark} reveals a substantial gap between text-only and image-based geological interpretation. For DeepSeek V4.1 Flash, providing images increases the Open score from 17.31 to 33.84 and the Overall score from 31.31 to 38.29, despite a decrease in MCQ from 52.30 to 44.96. This contrast suggests that multiple-choice performance can be partially supported by answer options and prior knowledge, whereas open-ended interpretation relies more heavily on visual evidence. Among the evaluated agents, GeoMM-Agent~\cite{geommbench2026} achieves an Overall score of 39.64, while all three \hygeoagent configurations outperform it.

Compared with their corresponding base models, all three \hygeoagent configurations improve across RQ1--RQ3. DeepSeek Agent achieves particularly substantial gains, increasing RQ1 from 20.04 to 28.26, RQ2 from 22.38 to 27.60, and RQ3 from 23.31 to 52.92, indicating improvements in hierarchical reasoning, prior--evidence adaptation, and cross-spectral interpretation. Its Overall score also rises from 38.29 to 46.23, representing a gain of 7.94 points. Claude Agent achieves the best overall performance, with an Overall score of 50.62, surpassing Claude Opus 5.5 (48.20) by 2.42 points. It also leads all evaluated configurations in MCQ (63.81), Open (41.82), and all three diagnostic metrics, achieving 34.63 on RQ1, 31.76 on RQ2, and 71.66 on RQ3.

Figure~\ref{fig:qualitative_comparison} presents representative cases across the three diagnostic dimensions. For Reasoning Reliability, models show varying abilities in describing target morphology and spatial patterns, while \hygeoagent uses region observation and coordinate grids to refine local judgments. For Prior--Evidence Contrast, some models fail to reconcile general geological knowledge with scene-specific evidence, whereas \hygeoagent combines geological retrieval and visual inspection to distinguish general expectations from image-supported conclusions. For Spectral Discrimination, models exhibit difficulties in jointly interpreting multiple band composites and identifying geological units, while \hygeoagent combines geological retrieval, spatial inspection, and reference comparisons to support multi-view interpretation. These examples illustrate how adaptive tool use helps address different sources of interpretation errors through complementary geological and visual information.

Table~\ref{tab:ablation_study} compares eight configurations using the same DeepSeek V4.1 Flash backbone on a 100-question subset, assigning zero credit to unanswered questions. The Full Agent scores 51.05, compared with 45.85 for direct answering. Removing local image observation or text retrieval reduces the reported mean by 4.60 and 4.22 points, respectively. The text retrieval ablation yields a 10.00-point decrease on MCQ and a 0.38-point decrease on Open. The Full Agent also exceeds the static PCA and grayscale variants by 3.45 and 3.62 points, respectively.

\section{Conclusion}
\label{sec:conclusion}

In this paper, we presented \hygeo, a diagnostic benchmark for evaluating MLLMs in remote sensing lithology interpretation across hierarchical geological reasoning, prior--evidence contrast, and cross-band interpretation. Evaluations reveal limitations in fine-grained geological recognition, evidence grounding, and cross-band consistency. To address these challenges, we proposed \hygeoagent, which integrates local visual observation, geological knowledge retrieval, same-band image retrieval, and dynamic multiband PCA analysis through adaptive tool use. Experiments demonstrate consistent improvements over base models, while ablation studies validate the contributions of these complementary tools. Future work will explore broader geological coverage, multisource observations, and more reliable evidence verification to further advance multimodal geological interpretation.

\section*{LLM Usage Statement}

ChatGPT 6.1 Sol was used during dataset construction
to assist in generating questions, multiple-choice options, and
open-ended questions. In addition, several large language models
(LLMs), including ChatGPT 6.1 Sol, were evaluated as experimental
models in this study. Their use in this work involved dataset
generation and model evaluation. All experimental results and
conclusions are the responsibility of the authors.

{
    \small
    \bibliographystyle{ieeenat_fullname}
    \bibliography{main}
}

% WARNING: do not forget to delete the supplementary pages from your submission 
% \clearpage
% \input{sec/X_suppl}

\end{document}